\documentclass[letterpaper, 10 pt, conference]{ieeeconf}  %
\usepackage{times}
\usepackage{luatex85}

\usepackage[bookmarks=true]{hyperref}

\usepackage[keeplastbox]{flushend}

\usepackage{booktabs}       %
\usepackage{amsfonts,amssymb,mathtools}       %

\usepackage{subcaption}
\usepackage{multirow}

\usepackage{algorithm}
\usepackage{algorithmicx}
\usepackage[noend]{algpseudocode}
\usepackage{cleveref}
\usepackage{acronym}
\usepackage{verbatim}
\usepackage{siunitx}

\crefname{appsec}{Appendix}{Appendices}

\usepackage{todonotes}
\usepackage{soul}
\definecolor{smoothgreen}{rgb}{0.7,1,0.7}
\sethlcolor{smoothgreen}

\usepackage[utf8]{inputenc}
\RequirePackage{luatex85}
\usepackage{pgfplots}
\pgfplotsset{compat=newest}
\pgfplotsset{every axis legend/.append style={%
cells={anchor=west}}
}
\usepgfplotslibrary{polar}
\usetikzlibrary{arrows}
\tikzset{>=stealth'}

\definecolor{C1}{rgb}{0.0, 0.447, 0.741}
\definecolor{C1_light}{rgb}{0.0, 0.6032388663967612, 1.0}
\definecolor{C2}{rgb}{0.85, 0.325, 0.098}
\definecolor{C3}{rgb}{0.929, 0.694, 0.125}
\definecolor{C4}{rgb}{0.494, 0.184, 0.556}
\definecolor{C5}{rgb}{0.466, 0.674, 0.188}
\definecolor{C6}{rgb}{0.301, 0.745, 0.933}
\definecolor{C7}{rgb}{0.635, 0.078, 0.184}

\definecolor{nice-red}{HTML}{E41A1C}
\definecolor{nice-orange}{HTML}{FF7F00}
\definecolor{nice-yellow}{HTML}{FFC020}
\definecolor{nice-green}{HTML}{4DAF4A}
\definecolor{nice-blue}{HTML}{377EB8}
\definecolor{nice-nice-red}{HTML}{984EA3}
\usepgfplotslibrary{groupplots}

\usetikzlibrary{shapes.geometric, arrows}

\tikzstyle{startstop} = [rectangle, rounded corners, minimum width=2cm, minimum height=1cm,text centered, draw=black, fill=none]
\tikzstyle{arrow} = [thick,->,>=stealth]

\usetikzlibrary{backgrounds}

\usepackage{ifthen}

\algnewcommand{\LineComment}[1]{\State\(\triangleright\) #1}

\IEEEoverridecommandlockouts                              %

\title{\LARGE \bf
EnsembleDAgger: A Bayesian Approach \\ to Safe Imitation Learning
}

\author{Kunal Menda,$^{1}$ Katherine Driggs-Campbell,$^{2}$ and Mykel J. Kochenderfer$^{1}$
	\thanks{$^{1}$Kunal Menda and Mykel J. Kochenderfer are at Stanford University, Stanford, CA 94305, USA
        {\tt\scriptsize \{kmenda,mykel\}@stanford.edu}}
    \thanks{$^{2}$Katherine Driggs-Campbell is at the University of Illinois at Urbana-Champaign, IL 61820, USA
        {\tt\scriptsize krdc@illinois.edu}}
}

\renewcommand{\baselinestretch}{0.98}

\begin{document}

\maketitle
\thispagestyle{plain}
\pagestyle{plain}

\begin{abstract}
Although imitation learning is often used in robotics, the approach frequently suffers from data mismatch and compounding errors.
DAgger is an iterative algorithm that addresses these issues by aggregating training data from both the expert and novice policies, but does not consider the impact of safety. We present a probabilistic extension to DAgger, which attempts to quantify the confidence of the novice policy as a proxy for safety.
Our method, EnsembleDAgger, approximates a Gaussian Process using an ensemble of neural networks.
Using the variance as a measure of confidence, we compute a decision rule that captures how much we doubt the novice, thus determining when it is safe to allow the novice to act.
With this approach, we aim to maximize the novice's share of actions, while constraining the probability of failure.
We demonstrate improved safety and learning performance compared to other DAgger variants and classic imitation learning on an inverted pendulum and in the MuJoCo HalfCheetah environment.
\end{abstract}

\section{Introduction}

To be truly intelligent, robotic systems must have the ability to learn by exploring their environment and state space in a safe way \cite{amodei2016concrete}.
One method to guide exploration is to learn from expert demonstrations~\cite{price2003accelerating,schaal1997learning,Kober2010}.
In contrast to reinforcement learning, where an explicit reward function must be defined, imitation learning guides exploration through expert supervision, allowing a robot to effectively learn from direct experience~\cite{ARGALL2009469}.
However, such supervised approaches are often suboptimal or fail when the policy that is being trained (referred to as the novice policy) encounters situations that are not adequately represented in the dataset provided by the expert \cite{daume2009search,ross2010efficient}.
While failures may be insignificant in simulation, safe learning is important when acting in the real world~\cite{amodei2016concrete}.

There are several methods for guided policy search in imitation learning settings~\cite{levine2013guided}.
One example is DAgger, which improves the training dataset by aggregating new data from both the expert and novice policies~\cite{ross2010efficient}.
DAgger has many desirable properties, including online functionality and theoretical guarantees.
This approach, however, does not guarantee safety.
Recent work extended DAgger to address some inherent drawbacks~\cite{kim2013maximum,laskey2016shiv}.
In particular, SafeDAgger augments DAgger with a decision rule policy to provide safe exploration while minimizing queries to the expert~\cite{Zhang2016}.

The shared goal of these methods is to efficiently train the novice to control the system while minimizing expert intervention.
These algorithms assume that by allowing the novice to act, the system will likely deviate from the expert trajectory set and sample a new state.
There is a chance, however, that the state visited is unsafe, or is a failure state. %
If the expert acts instead, we assume that the system will move along a safe trajectory, which is likely through states similar to those previously observed.
The goal of this paper is to present an algorithm that maximizes the novice's share of actions, while constraining the probability of failure. %

\begin{figure*}[t!]
	\centering
	\includegraphics[width=0.97\textwidth]{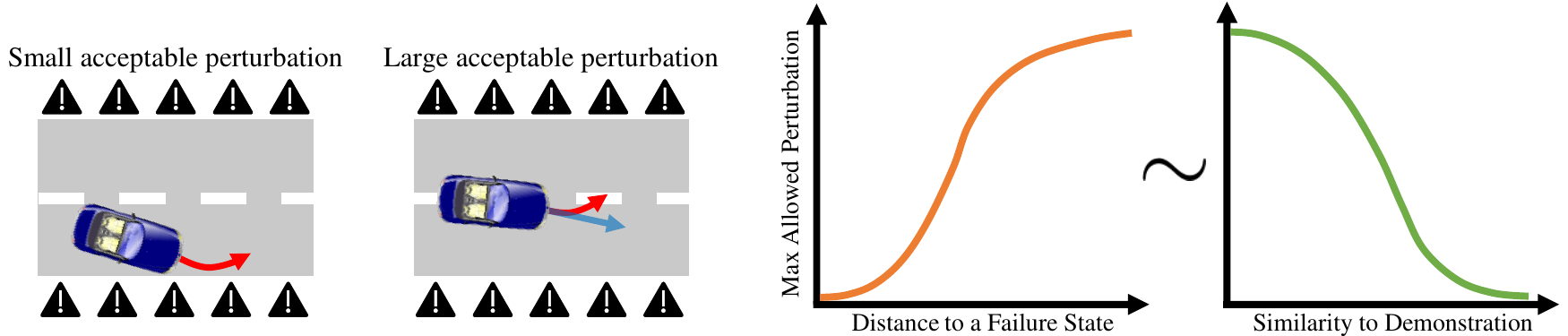}
	\caption{\small Visualization of the tradeoffs between familiarity and risk. (left) Example scenarios of where perturbations are (not) permissible due to low (high) risk.  Red trajectories illustrate expert corrections and the blue trajectory illustrate novice actions. (right) Plots visualizing the ideal tradeoff between distance to failure state and allowed deviations and the approximate of this tradeoff using similarity to expert demonstrations and deviations.}
	\label{fig:intro}
\end{figure*}

Ideally, the proximity to a failure state (measured as an $l_2$-distance or likelihood of encountering the state under some operating condition) is known, and a safety envelope can be computed to guarantee safety \cite{akametalu2014reachability}.
In the case of model-free learning, such guarantees are much more difficult to make.
If we consider the novice action to be a perturbed form of the expert action, then we hypothesize that for many systems, the magnitude of permissible perturbation to expert actions is related to the distance from unsafe regions.
Further, in a model-free case where expert demonstrations are available, we hypothesize that there is an inverse relationship between a state's similarity to those in expert trajectories and allowed perturbations. 
We visualize this intuition in \Cref{fig:intro}. 
In the left panel, we see that the maximum permissible deviation from an expert action should be low as the system approaches a wall, which is considered a dangerous state. 
In such settings, experts will likely prefer trajectories that maintain some margin of distance from unsafe states. 
Assuming this to be the case, it follows that in unfamiliar states, the system is likely at higher risk of entering failure states, and thus it is safer to allow the expert to act.
While in familiar regions, it is permissible for the novice to act with large deviation from expert action.

This paper extends DAgger to a probabilistic domain, and aims to minimize expert intevention while constraining the likelihood of failure.
While SafeDAgger uses the \emph{discrepancy} between the expert and the novice to determine safety, we measure \emph{doubt} by quantifying the uncertainty or confidence of the novice policy. 
To quantify doubt, we use an ensemble of neural networks to estimate the variance of the novice action in a particular state, which we show can effectively approximate Gaussian Processes (GPs), even in complex, high-dimensional spaces~\cite{rasmussen2004gaussian}.

We demonstrate how our method out-performs existing DAgger variants in an imitation learning setting. 
This paper makes two key contributions: 
(1) we present EnsembleDAgger, a Bayesian extension to DAgger, which introduces a probabilistic notion of safety to minimize expert intervention while constraining the probability of failure; and
(2) we demonstrate the utility of this approach with improved performance and safety in an imitation learning case study on an inverted pendulum and demonstrate the scalability of the approach on the MuJoCo HalfCheetah domain.

\section{Background}
\label{sec:related_works}

This section presents a brief technical overview of DAgger, SafeDAgger, and different methods for approximating GPs using neural networks.

\subsection{DAgger and SafeDAgger}
The DAgger framework extends traditional supervised learning approaches by simultaneously running both an expert policy that we wish to clone and a novice policy we wish to train \cite{ross2011reduction}.
By aggregating new data from the expert, the underlying model and reward structure are uncovered.

Using supervised learning, we train an initial novice policy $\pi_{\text{nov},0}$ on some initial training set $\mathcal{D}_0$ generated by the expert policy $\pi_{\text{exp}}$. 
With this initialization, DAgger iteratively collects additional training examples from a mixture of the expert and novice policy. During a given episode, the combined expert and novice system interacts with the environment under the supervision of a decision rule. 
The decision rule, referred to as DR$(\cdot)$ in \Cref{alg:dagger}, decides at every time-step $t$ whether to use the action from the novice or expert to interact with the environment (\Cref{fig:combinedsystemflowchart}). 
The observations $o_t$ received during each epoch %
and the expert's choice of corresponding actions make up a new dataset called $\mathcal{D}_i$. 
The new dataset of training examples is combined with the previous sets: $\mathcal{D} \gets \mathcal{D} \cup \mathcal{D}_i$, and the novice policy is then re-trained on $\mathcal{D}$, as presented in \Cref{alg:dagger}.

\begin{figure}[!t]
	\centering
	\noindent
	\begin{minipage}{0.46\textwidth}
		\centering
		\scalebox{0.8}{\begin{tikzpicture}[node distance=1cm]

\node (piexp) [startstop] {$\pi_\text{exp}(o_t)$};
\node (pinov) [startstop, below of=piexp, yshift=-0.5cm] {$\pi_\text{nov}(o_t)$};
\node (decisionsystem) [startstop, right of=piexp, xshift=2.5cm, yshift=-0.5cm] {Decision Rule};
\node (env) [startstop, below of=pinov, xshift=1.5cm, yshift=-0.5cm] {Environment};

\draw [arrow] (piexp.east) -- node[anchor=south] {$a_{\text{exp},t}$} ++(+1.0,0) |- (decisionsystem.170);
\draw [arrow] (pinov.east) -- node[anchor=north] {$a_{\text{nov},t}$} ++(+1.0,0) |- (decisionsystem.190);
\draw [arrow] (decisionsystem.east) |- node[anchor=south west] {$a_t$} ++(+0.5,0) |- (env.east);
\draw [arrow] (env.west) -- node[anchor=north] {$o_t$} ++(-2,0) |- (piexp.west);
\draw [arrow] (env.west) -- ++(-2,0) |- (pinov.west);

\end{tikzpicture}}
		\caption{\small Flowchart for DAgger variants, where the decision rule differs between approaches.
			\label{fig:combinedsystemflowchart}}
	\end{minipage} \quad
	\begin{minipage}{0.5\textwidth}
		\centering
		\begin{algorithm}[H]
			\caption{\textsc{DAgger} \label{alg:dagger}}
			\begin{algorithmic}[1]
				\Procedure{DAgger}{DR($\cdot$)}
				\State Initialize $\mathcal{D} \gets \emptyset $
				\State Initialize $\pi_{\text{nov},i}$
				\For{epoch $i=1:K$}
				\State Sample $T$-step trajectories using $\text{DR}$ 
				\State Get $\mathcal{D}_i = \left\{o_t,\pi_\text{exp}(o_t)\mid t\in1:T\right\}$
				\State Aggregate datasets: $\mathcal{D} \gets \mathcal{D}\cup\mathcal{D}_i$
				\State Train $\pi_{\text{nov},i+1}$ on $\mathcal{D}$
				\EndFor
				\EndProcedure
			\end{algorithmic}
		\end{algorithm}
	\end{minipage}
\end{figure}

By allowing the novice to act, the combined system explores parts of the state space further from the nominal trajectories of the expert. 
In querying the expert in these parts of the state space, the novice is able to learn a more robust policy. 
However, allowing the novice to always act risks the possibility of encountering an unsafe state, which can be costly in real-world experiments. 
The VanillaDAgger algorithm and SafeDAgger balance this trade-off by their choice of decision rules. 

\begin{figure}[!t]
	\centering
	\noindent
	\begin{minipage}{\linewidth}
		\begin{algorithm}[H]
			\caption{\textsc{VanillaDAgger} Decision Rule\label{alg:vanilla_dagger_decision_rule}}
			\begin{algorithmic}[1]
				\Procedure{DR}{$o_t, i, \beta_0, \lambda$}
				\State $a_{\text{nov}, t} \gets \pi_{\text{nov}}(o_t)$
				\State $a_{\text{exp}, t} \gets \pi_{\text{exp}}(o_t)$
				\State $\beta_i \gets \lambda^i\beta_0$
				\State $ z \sim \text{Uniform}(0,1)$
				\If{$z \leq \beta_i$}
				\State \textbf{return} $a_{\text{exp}, t}$
				\Else
				\State \textbf{return} $a_{\text{nov}, t}$
				\EndIf
				\EndProcedure
			\end{algorithmic}
		\end{algorithm}
		\begin{algorithm}[H]
			\caption{\textsc{SafeDAgger*} Decision Rule\label{alg:safe_dagger_decision_rule}}
			\begin{algorithmic}[1]
				\Procedure{DR}{$o_t, \tau$}
				\State $a_{\text{nov}, t} \gets \pi_{\text{nov}}(o_t)$
				\State $a_{\text{exp}, t} \gets \pi_{\text{exp}}(o_t)$
				\If{$\lVert a_{\text{nov}, t} - a_{\text{exp}, t}\rVert^2 \leq \tau$}
				\State \textbf{return} $a_{\text{nov}, t}$
				\Else
				\State \textbf{return} $a_{\text{exp}, t}$
				\EndIf
				\EndProcedure
			\end{algorithmic}
		\end{algorithm}
	\end{minipage}
\end{figure}

Under VanillaDAgger (\Cref{alg:vanilla_dagger_decision_rule}), the expert's action is chosen with probability $\beta_i\in [0,1]$, where $i$ denotes the DAgger epoch. 
If $\beta_i = \lambda \beta_{i-1}$ for some $\lambda\in(0,1)$, then the novice takes increasingly more actions each epoch. As the novice is given more training labels from previous epochs, it is allowed greater autonomy in exploring the state space. 
The VanillaDAgger decision-rule does not consider any similarity measure between the novice and expert actions. Hence, even if the novice suggests a highly unsafe action, VanillaDAgger allows the novice to act with probability $(1-\beta_i)$. 

The ``optimal'' decision-rule approximated by SafeDAgger, presented in \Cref{alg:safe_dagger_decision_rule} and referred to as SafeDAgger*, computes the \emph{discrepancy} between the expert and novice actions and allows the novice to act if the distance between the actions is less than some chosen threshold~$\tau$~\cite{Zhang2016}.\footnote{To reduce the number of expert queries, SafeDAgger approximates the SafeDAgger* decision rule using a deep policy that determines whether or not the novice policy is likely to deviate from the reference policy. Unlike SafeDAgger, we are not concerned with minimizing expert queries during a given episode. Hence, we compare to the SafeDAgger* decision rule directly, as opposed to the approximation. } Though this decision rule is claimed to be optimal, we argue that it has a shortcoming. 

An ideal decision rule would allow the novice to act if there is a sufficiently low probability that the system can transition to an unsafe state. If the combined system is currently near an unsafe state, the tolerable perturbation from the expert's choice of action is smaller than when the system is far from unsafe states. Hence, in practice, the single threshold $\tau$ employed in SafeDAgger* is either too conservative when the system is far from unsafe states or too relaxed when near them.
To approximate the ideal decision rule in a model-free manner, we propose not just considering the distance between the novice's and expert's actions, but also the uncertainty in the novice policy at a given state. To estimate the uncertainty of the novice policy, we use Bayesian deep learning.

There are two works which build on the algorithm presented in this work. Kelly et al.~\cite{kelly2018hg} perform experiments on an autonomous vehicle and find a safe method to query humans for demonstrations, and calibrating the threshold parameters of the algorithm presented in our work. Cronrath et al.~\cite{cronrathbagger} propose an extension of our ideas that attempt to combine the improved safety of a Bayesian extension to DAgger with the query efficiency of SafeDAgger.

\subsection{Bayesian Approximation Methods}
\label{sec:bayesappx}

Recent research has focused on approximating GPs with neural networks \cite{lee2017deep}.
While GPs alone have shown great success in modeling uncertainty and approximating safety \cite{berkenkamp2016safe}, traditional GP approaches are computationally expensive for high-dimensional feature spaces and large datasets \cite{rasmussen2004gaussian}.
Advances in deep learning have shown great success in handling these complexities.
Two methods for approximating GPs with deep neural networks are ensemble methods \cite{lakshminarayanan2017simple} and Monte-Carlo dropout~\cite{Gal2015DropoutB}. 
Refer to~\Cref{app:GP_approx} for a summary of advantages and disadvantages of these approaches and an empirical evaluation of these methods.

In this work, we chose to use the ensemble method, which is a technique for training a collection of neural networks to execute the same task and then combining the output into a single prediction.
This approach has shown to significantly improve performance in practice~\cite{zhou2002ensembling}. 
There is a work that employed an ensemble of neural networks to approximate GPs and demonstrated that this is a more straightforward approach to estimate predictive uncertainty~\cite{lakshminarayanan2017simple}. 
Typically, neural networks predict point estimates of the output that are optimized to minimize the mean squared error on the training set. 
The authors claim that this approach does not capture irreducible, or \textit{aleatoric} uncertainty, but only \textit{epistemic} uncertainty. They propose using a \emph{proper scoring rule}, like negative log-likelihood, as a loss function to train an ensemble in which each network predicts a mean and a variance of a Gaussian distribution over the output. They postulate that such loss functions provide a better measure of the quality of predictive uncertainty and thus reward better calibrated predictions. Network predictions are then combined as a mixture of Gaussians.

\section{EnsembleDAgger}
\label{sec:methods}

We present the EnsembleDAgger decision rule, in which the \textit{discrepancy} between the expert's and the novice's mean action, as well as the novice's \textit{doubt}, which is variance of the novice's action, are used to decide whether to choose the novice action. 
According to the EnsembleDAgger decision rule, the novice must satisfy two conditions in order to act. The first is that the discrepancy between the novice and expert's action, i.e. $\lVert \bar{a}_{\text{nov}, t} - a_{\text{exp}, t}\rVert^2$, must be less than some threshold $\tau$. 
This is the SafeDAgger* decision rule, but will henceforth be referred to as the \textit{discrepancy rule}.
Assuming the novice policy outputs a variance on its predicted action $\sigma^2_{a_{\text{nov}, t}}$, as an ensemble of neural networks would, then the second condition is that $\sigma^2_{a_{\text{nov}, t}}$ is less than some threshold $\chi$. 
We refer this condition as the \textit{doubt rule}. 
As shown in \Cref{fig:ed_decrule}, in order for the novice to act according to the EnsembleDAgger decision rule, it must satisfy both the discrepncy rule and the double rule. 
The algorithm, described in \Cref{alg:ensemble_decision_rule}, is parameterized by the values $\tau$ and $\chi$.

We restate the assumptions made to explain why the this decision rule is able to better guarantee the system's safety:
\begin{enumerate}%
    \item The expert prefers trajectories that avoid failure states, and rarely visits near failure states, implying that states dissimilar to those in expert trajectories (or states unfamiliar to the novice) are likely to be in closer proximity to failure states.
    \item Following from (1), and by capturing epistemic uncertainty, or lack of familiarity with states in the training dataset, the novice's doubt provides a model-free proxy for proximity to failure states.
    \item In order to constrain the probability of encountering a failure state, the discrepancy between the action taken and the expert's action is less than some bound.
    \item The ideal bounds should be state-dependent, such that the bound is tighter in close proximity to failure states. 
    \item Following from (2, 4), the bound on discrepancy should decrease as the novice's doubt increases.
\end{enumerate}

Also, it is assumed that the expert policy is primarily unimodal, as is commonly assumed in most imitation learning settings. 
Further, using a neural network based dissimilarity measure is useful for imitation learning as neural networks scale more gracefully to high-dimensional input spaces and large datasets than most non-parametric measures.

Given that we have a measure of doubt via the variance on novice actions, we ideally would like to specify the bound on discrepancy as a monotonically increasing function of doubt. 
To meet this end, we have experimented with the idea of making the discrepancy bound proportional to the inverse of doubt. 
However, the parameters specifying an arbitrary function mapping doubt to a discrepancy bound must be considered hyperparameters to the algorithm and tuned by the practitioner. 
We opt for the low-order approximation to the ideal functional mapping, shown in \Cref{fig:ed_decrule}, because the two hyperparameters, $\chi$ and $\tau$, are easy to interpret.

\begin{figure}[!t]
	\centering
	\noindent
	\begin{minipage}{0.48\textwidth}
		\centering
    \scalebox{0.9}{\begin{tikzpicture}[ultra thick]
\coordinate (tlc) at (0.0, 5.0);
\coordinate (blc) at (0.0, 0.0);
\coordinate (brc) at (6.0, 0.0);

\coordinate (bm) at (3.0, 0.0);
\coordinate (tm) at (3.0, 4.75);
\coordinate (lm) at (0.0, 2.5);
\coordinate (rm) at (5.75, 2.5);

\coordinate (ct) at (0.05, 4.5);
\coordinate (mp) at (3.0, 2.5);
\coordinate (cb) at (5.5, 0.05);

\node at (3.0, -0.25) {Novice Doubt};
\node[rotate=90] at (-0.25, 2.5) {Novice Discrepancy};
\node at (1.5,1.25) {Novice Acts};
\node at (3.0, 5.0) {\large $\chi$};
\node at (6.0, 2.5) {\large $\tau$};

\draw [black] (tlc) -- (blc);
\draw [black] (blc) -- (brc);

\draw [C5]  (ct) to[out=-10,in=110] (mp);
\draw [C5]  (mp) to[out=-70,in=170] (cb);

\draw [C2, dashed] (bm) -- (tm);
\draw [C2, dashed] (lm) -- (rm);

\end{tikzpicture}}
		\caption{\small The EnsembleDAgger decision rule is parametrized by doubt ($\chi$) and discrepancy ($\tau$) bounds, and is a low-order, model-free approximation to the `ideal' decision rule, shown in green.}
		\label{fig:ed_decrule}
	\end{minipage} \quad

	\begin{minipage}{0.48\textwidth}
	\begin{algorithm}[H]
		\caption{EnsembleDAgger Decision Rule\label{alg:ensemble_decision_rule}}
		\begin{algorithmic}[1]
			\Procedure{DR}{$o_t, \tau, \chi$}
			\State $\bar{a}_{\text{nov}, t}, \sigma^2_{a_{\text{nov}, t}} \gets \pi_{\text{nov}}(o_t)$ %
			\State $a_{\text{exp}, t} \gets \pi_{\text{exp}}(o_t)$
			\State $\hat{\tau} \gets \lVert \bar{a}_{\text{nov}, t} - a_{\text{exp}, t}\rVert^2 $
			\State $\hat{\chi}  \gets  \sigma^2_{a_{\text{nov}, t}} $
			\If{$\hat{\tau} \leq \tau$ \textbf{and} $\hat{\chi} \leq \chi$}
			\State \textbf{return} $\bar{a}_{\text{nov}, t}$
			\Else
			\State \textbf{return} $a_{\text{exp},t}$
			\EndIf
			\EndProcedure
		\end{algorithmic}
	\end{algorithm}
	\end{minipage}
\end{figure}

By appropriately choosing the hyper-parameters $\tau$ and $\chi$, we satisfy the dual objectives of allowing the novice to act only if it is sufficiently confident in its action and close to the expert. 
As $\chi \rightarrow \infty$, the decision rule converges to that of SafeDAgger*. As $\tau \rightarrow \infty$, the decision rule ignores discrepancy, and allows the novice to act if it is confident without comparison to what the expert action is. However, since the novice is only confident in states similar to those in $\mathcal{D}$, it is likely that the novice having low doubt causes its action to also have low discrepancy, implying that the algorithm is less sensitive to an arbitrary increase in $\tau$ than to an arbitrary increase in $\chi$. This statement is qualified in the next section by showing that using the doubt rule alone (by setting $\tau = \infty$) leads to better performance than using the discrepancy rule alone (by setting $\chi = \infty$).
Though not the focus of this work, it is also worth noting that the expert only needs to be queried if the doubt rule is satisfied, thereby leading to query efficiency.

\section{Experiments}
\label{sec:experiments}

In this section, we present experimental validation for the following claims we have made:
\begin{itemize}
	\item Using the discrepancy rule alone with fixed $\tau$ is wastefully conservative in some regions of the state space, while not conservative enough in others.
	\item The variance of the novice policy's output is a good measure of dissimilarity between the query state and states in the training dataset.
	\item Using the doubt rule alone with fixed $\chi$ trains a better performing novice policy for the same compromise to the combined (expert and novice) system's safety.
	\item Combining the two decision rules in the EnsembleDAgger framework improves the trained novice policy performance while making the combined system strictly safer.
\end{itemize}

To justify the first two of the above claims, we make use of a simple inverted pendulum domain. Such a simple domain is chosen because it allows us to visualize the portion of the state-space in which a given decision rule allows a given novice policy to act. We justify the latter two claims on the MuJoCo HalfCheetah OpenAI Gym environment. 

\subsection{Inverted Pendulum domain}

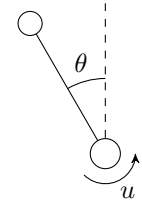
\begin{figure}[t]
	\centering
		\begin{tikzpicture}[ scale=2]
		\draw (0,0) -- (120:1);
		\draw[dashed] (0,0) -- (0,1);
		\draw (90:0.5) arc (90:120:0.5);
		\draw[fill=white] (0,0) circle (0.1);
		\draw[fill=white] (120:1) circle (0.08);
		\node (th) at (105:0.625) {$\theta$};
		\draw[->] (225:0.2) arc (225:360:0.2);
		\node (u) at (300:0.3) {$u$};
		\end{tikzpicture}
		\caption{\small The inverted pendulum environment has a state space of $[\theta, \dot{\theta}]$ and an action space of the torque $u$. \label{fig:invpen_diagram}}
		\vspace{-10pt}
\end{figure}

    \begin{figure}[t!]
        \centering
		\scalebox{0.6}
		{\begin{tikzpicture}[]
\begin{axis}[ylabel = {$\dot{\theta}$ [rad/s]}, xmin = {-3.14}, xmax = {3.14}, ymax = {4.999}, xlabel = {$\theta$ [rad]}, grid=both, ymin = {-4.999}]\addplot+[draw=none, draw opacity=0, mark=*, mark options={fill=blue, fill opacity = 0.25}] coordinates {
(-3.1415927, -5.0)
(-2.8108988, -5.0)
(1.4881228, -5.0)
(1.8188168, -5.0)
(2.1495109, -5.0)
(2.4802048, -5.0)
(2.8108988, -5.0)
(3.1415927, -5.0)
(-3.1415927, -4.4736843)
(1.1574289, -4.4736843)
(1.4881228, -4.4736843)
(1.8188168, -4.4736843)
(2.1495109, -4.4736843)
(2.4802048, -4.4736843)
(2.8108988, -4.4736843)
(3.1415927, -4.4736843)
(-3.1415927, -3.9473684)
(0.8267349, -3.9473684)
(1.1574289, -3.9473684)
(1.4881228, -3.9473684)
(1.8188168, -3.9473684)
(2.1495109, -3.9473684)
(2.4802048, -3.9473684)
(2.8108988, -3.9473684)
(3.1415927, -3.9473684)
(-3.1415927, -3.4210527)
(0.49604094, -3.4210527)
(0.8267349, -3.4210527)
(1.1574289, -3.4210527)
(1.4881228, -3.4210527)
(1.8188168, -3.4210527)
(2.1495109, -3.4210527)
(2.4802048, -3.4210527)
(2.8108988, -3.4210527)
(0.16534698, -2.8947368)
(0.49604094, -2.8947368)
(0.8267349, -2.8947368)
(1.1574289, -2.8947368)
(1.4881228, -2.8947368)
(1.8188168, -2.8947368)
(2.1495109, -2.8947368)
(2.4802048, -2.8947368)
(2.8108988, -2.8947368)
(-0.16534698, -2.368421)
(0.16534698, -2.368421)
(0.49604094, -2.368421)
(0.8267349, -2.368421)
(1.1574289, -2.368421)
(1.4881228, -2.368421)
(1.8188168, -2.368421)
(2.1495109, -2.368421)
(2.4802048, -2.368421)
(-0.49604094, -1.8421053)
(-0.16534698, -1.8421053)
(0.16534698, -1.8421053)
(0.49604094, -1.8421053)
(0.8267349, -1.8421053)
(1.1574289, -1.8421053)
(1.4881228, -1.8421053)
(1.8188168, -1.8421053)
(2.1495109, -1.8421053)
(-0.8267349, -1.3157895)
(-0.49604094, -1.3157895)
(-0.16534698, -1.3157895)
(0.16534698, -1.3157895)
(0.49604094, -1.3157895)
(0.8267349, -1.3157895)
(1.1574289, -1.3157895)
(1.4881228, -1.3157895)
(1.8188168, -1.3157895)
(2.1495109, -1.3157895)
(-1.1574289, -0.7894737)
(-0.8267349, -0.7894737)
(-0.49604094, -0.7894737)
(-0.16534698, -0.7894737)
(0.16534698, -0.7894737)
(0.49604094, -0.7894737)
(0.8267349, -0.7894737)
(1.1574289, -0.7894737)
(1.4881228, -0.7894737)
(1.8188168, -0.7894737)
(-1.1574289, -0.2631579)
(-0.8267349, -0.2631579)
(-0.49604094, -0.2631579)
(-0.16534698, -0.2631579)
(0.16534698, -0.2631579)
(0.49604094, -0.2631579)
(0.8267349, -0.2631579)
(1.1574289, -0.2631579)
(1.4881228, -0.2631579)
(-1.4881228, 0.2631579)
(-1.1574289, 0.2631579)
(-0.8267349, 0.2631579)
(-0.49604094, 0.2631579)
(-0.16534698, 0.2631579)
(0.16534698, 0.2631579)
(0.49604094, 0.2631579)
(0.8267349, 0.2631579)
(1.1574289, 0.2631579)
(-1.8188168, 0.7894737)
(-1.4881228, 0.7894737)
(-1.1574289, 0.7894737)
(-0.8267349, 0.7894737)
(-0.49604094, 0.7894737)
(-0.16534698, 0.7894737)
(0.16534698, 0.7894737)
(0.49604094, 0.7894737)
(0.8267349, 0.7894737)
(1.1574289, 0.7894737)
(-2.1495109, 1.3157895)
(-1.8188168, 1.3157895)
(-1.4881228, 1.3157895)
(-1.1574289, 1.3157895)
(-0.8267349, 1.3157895)
(-0.49604094, 1.3157895)
(-0.16534698, 1.3157895)
(0.16534698, 1.3157895)
(0.49604094, 1.3157895)
(0.8267349, 1.3157895)
(-2.1495109, 1.8421053)
(-1.8188168, 1.8421053)
(-1.4881228, 1.8421053)
(-1.1574289, 1.8421053)
(-0.8267349, 1.8421053)
(-0.49604094, 1.8421053)
(-0.16534698, 1.8421053)
(0.16534698, 1.8421053)
(0.49604094, 1.8421053)
(-2.4802048, 2.368421)
(-2.1495109, 2.368421)
(-1.8188168, 2.368421)
(-1.4881228, 2.368421)
(-1.1574289, 2.368421)
(-0.8267349, 2.368421)
(-0.49604094, 2.368421)
(-0.16534698, 2.368421)
(0.16534698, 2.368421)
(-2.8108988, 2.8947368)
(-2.4802048, 2.8947368)
(-2.1495109, 2.8947368)
(-1.8188168, 2.8947368)
(-1.4881228, 2.8947368)
(-1.1574289, 2.8947368)
(-0.8267349, 2.8947368)
(-0.49604094, 2.8947368)
(-0.16534698, 2.8947368)
(-2.8108988, 3.4210527)
(-2.4802048, 3.4210527)
(-2.1495109, 3.4210527)
(-1.8188168, 3.4210527)
(-1.4881228, 3.4210527)
(-1.1574289, 3.4210527)
(-0.8267349, 3.4210527)
(-0.49604094, 3.4210527)
(3.1415927, 3.4210527)
(-3.1415927, 3.9473684)
(-2.8108988, 3.9473684)
(-2.4802048, 3.9473684)
(-2.1495109, 3.9473684)
(-1.8188168, 3.9473684)
(-1.4881228, 3.9473684)
(-1.1574289, 3.9473684)
(-0.8267349, 3.9473684)
(3.1415927, 3.9473684)
(-3.1415927, 4.4736843)
(-2.8108988, 4.4736843)
(-2.4802048, 4.4736843)
(-2.1495109, 4.4736843)
(-1.8188168, 4.4736843)
(-1.4881228, 4.4736843)
(-1.1574289, 4.4736843)
(3.1415927, 4.4736843)
(-3.1415927, 5.0)
(-2.8108988, 5.0)
(-2.4802048, 5.0)
(-2.1495109, 5.0)
(-1.8188168, 5.0)
(-1.4881228, 5.0)
(2.8108988, 5.0)
(3.1415927, 5.0)
};
\addlegendentry{Expert converages}
\addplot+ [mark = {none}, red, fill=red, fill opacity=0.3]coordinates {
(0.17444444444444446, -0.5)
(-0.17444444444444446, -0.5)
(-0.17444444444444446, 0.5)
(0.17444444444444446, 0.5)
(0.17444444444444446, -0.5)
};
\addlegendentry{Initial conditions}
\end{axis}

\end{tikzpicture}}
		\caption{\small This figure shows states in the expert's basin of attraction, i.e. states from which the expert converges to the origin. The figure also shows the set from which initial conditions of DAgger epochs are uniformly drawn in this experiment. \label{fig:dom_attraction}}
	\end{figure}
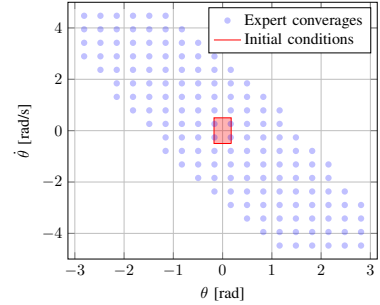

Following the experimental protocol presented by Berkenkamp et al., we concretely visualize behavior by considering a deterministic but non-linear control problem of stabilizing an inverted pendulum, which has a two-dimensional state space of $[\theta,\dot{\theta}]$ and a one-dimensional action space of $u$, as shown in~\Cref{fig:invpen_diagram}~\cite{berkenkamp2017safe}.
The control law was derived using by feedback linearization~\cite{khalilnonlinear}. 
\Cref{fig:dom_attraction} shows the controller's basin of attraction and highlights the states from which initial conditions are sampled uniformly during the successive epochs of DAgger.
The dynamics and control law are provided in~\Cref{app:invpen}.

The neural network model representing the novice policy is an ensemble of ten multi-layer perceptrons, each with four hidden layers of size $[64, 64, 32, 32]$ respectively. At each DAgger epoch, the ten networks are each trained for 200 training epochs with a learning rate of $10^{-3}$, $l_2$-weight regularization of $10^{-5}$, and a mini-batch size of 16. The maximum length of any trajectory %
is 100. No dropout or batch normalization is used. Since the data labeled by the deterministic expert is noise-free, the networks do not individually predict variance and are trained with MSE loss.

\begin{figure*}[!t]
	\centering
	\scalebox{0.62}
	{\input{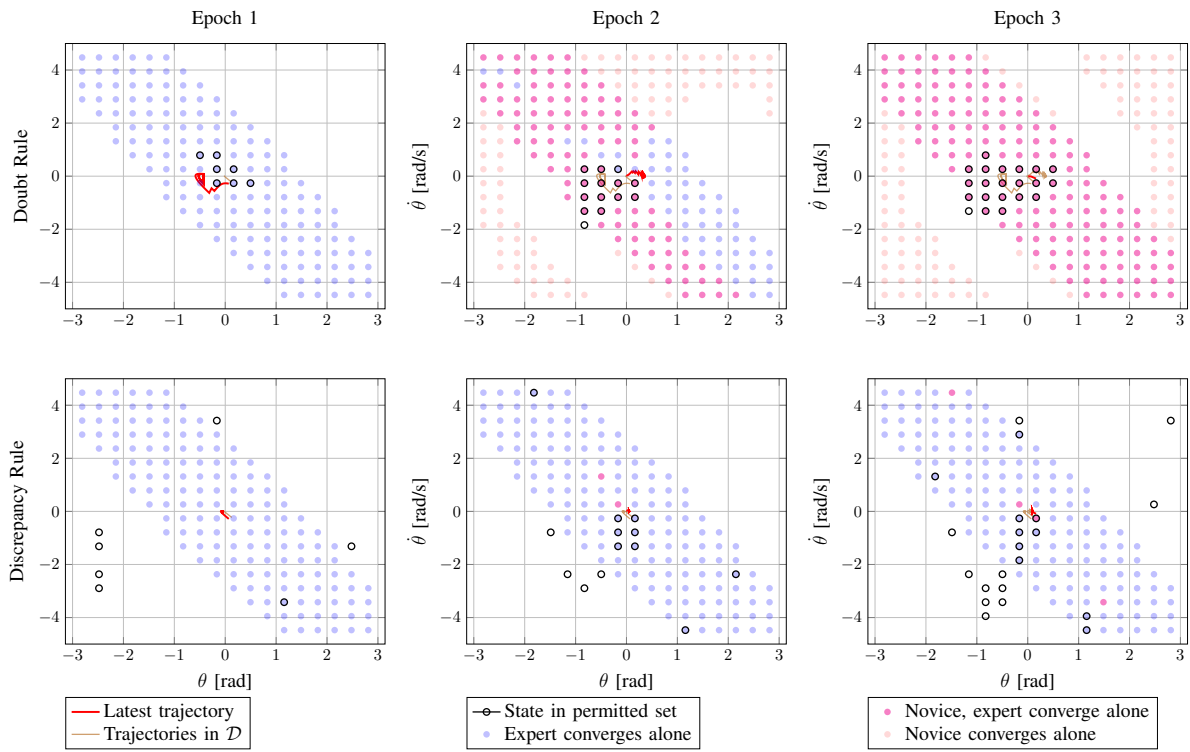}}
	\caption{\small Three epochs of DAgger are compared when using the doubt rule (top) and discrepancy rule (bottom). %
		The permitted set, i.e. states at which the novice is allowed to act, denoted by black circles, of the doubt rule is concentrated in the neighborhood of states represented in $\mathcal{D}$, where as the permitted set of the discrepancy rule is distributed more haphazardly across the state space. %
		States in which the novice alone is able to converge to the origin are indicated in pink. 
	}
	\label{fig:invpen_experiment}
\end{figure*}

In order to compare the two decision rules, we are interested in analyzing the regions of the state space in which they allow the novice to act. We define the \textit{permitted set} for some decision rule, given some novice and expert policies, to be the set of states in which the decision rule chooses the novice action. 
In \Cref{fig:invpen_experiment}, states in the permitted set are shown as black circles.
Similarly, we define the  \textit{permitted set volume} to be the fraction of states grid-sampled in $\theta \in [-\pi,\pi], \dot{\theta} \in [-5,5]$ that are in the permitted set of a given decision rule, given some novice and expert policies. 
Additionally, we define the \textit{novice basin of attraction} to be the set of states $\mathcal{X}_0$ from which, if the novice is initialized in $\mathcal{X}_0$ and allowed to act alone (without the help of the expert), the novice converges to the origin. 

In order to make an apples-to-apples comparison between the two decision rules, we provide a \textit{budget}, and analyze how the two decisions utilize this budget. The budget chosen is a fixed volume for the permitted set. At each epoch, since the novice has learned from more data, we linearly grow the permitted set volume budget. Prior to each episode, we solve for the value of $\chi$ and $\tau$ that will make the doubt and discrepancy rules respectively yield permitted sets with the desired volume.
These values are found using bisection search.

The goal of this experiment is to show that, for some fixed volume permitted set, the doubt rule allocates that volume in the neighborhood of states represented in $\mathcal{D}$, justifying the claim that the novice's output variance is a good measure of dissimilarity between the query state and familiar states. Additionally, we show that the discrepancy rule haphazardly allocates volume to regions of the state space in which the novice and expert agree by chance, indicating that it is wastefully conservative in some regions of the state space, while not conservative enough in others.

In an additional experiment on this domain that can be found in the~\Cref{app:invpenhyp}, we compare the decision rules in a manner meaningful to a practitioner, by fixing the hyperparameters \textit{a priori} and keeping them fixed over all epochs. For both experiments on this domain, we control the random seed specifying the initial condition for each epoch such that it varies across epoch but is the same regardless of decision rule. The trajectory followed from that initial condition will, of course, depend on the decision rule. In all experiments, as in all variants of DAgger, we initialize $\mathcal{D}$ with a zeroth epoch where only the expert is queried for the action, and the decision rule is used from the first epoch onward~\cite{ross2011reduction}. 

\Cref{fig:invpen_experiment} shows the evolution of the permitted set under the doubt rule and the discrepancy rule for the first three epochs of an experiment. Under the doubt rule, the permitted set is concentrated in the neighborhood of the labeled states in $\mathcal{D}$. This is because the variance of the function fitting $\mathcal{D}$ grows as we move away from labeled states, so the permitted set is constrained to be within some neighborhood of labeled states under the doubt rule.
On the other hand, under the discrepancy rule, the permitted set is more haphazardly distributed over the state space with a smaller portion of the allotted volume being in the neighborhood of labeled states. We observe this because there exist arbitrary regions of the state space in which the function fitting the $\mathcal{D}$ happens to intersect the true control law purely by chance, leading to low discrepancy in these, often dangerous, regions.

We can see in \Cref{fig:invpen_experiment} that the trajectories resulting under the doubt rule carry the system to the edge of a familiar region of the state space, after which the expert is handed control to navigate unfamiliar regions. This behavior leads to a novice basin of attraction that is much larger than under the discrepancy rule, while no trajectories enter dangerous territory. However, under the discrepancy rule, we see that the novice is rarely allowed to carry the system away from an expert trajectory, thereby aggregating a dataset that is not much more likely to be informative than behavior cloning.  This observation qualitatively suggests that the doubt rule can train a better performing novice policy for the same level of compromise to the combined (expert and novice) system's safety. This claim will be justified in the next experiment.

\subsection{MuJoCo HalfCheetah domain}
\begin{figure}
    \centering
    \includegraphics[width=0.5\columnwidth]{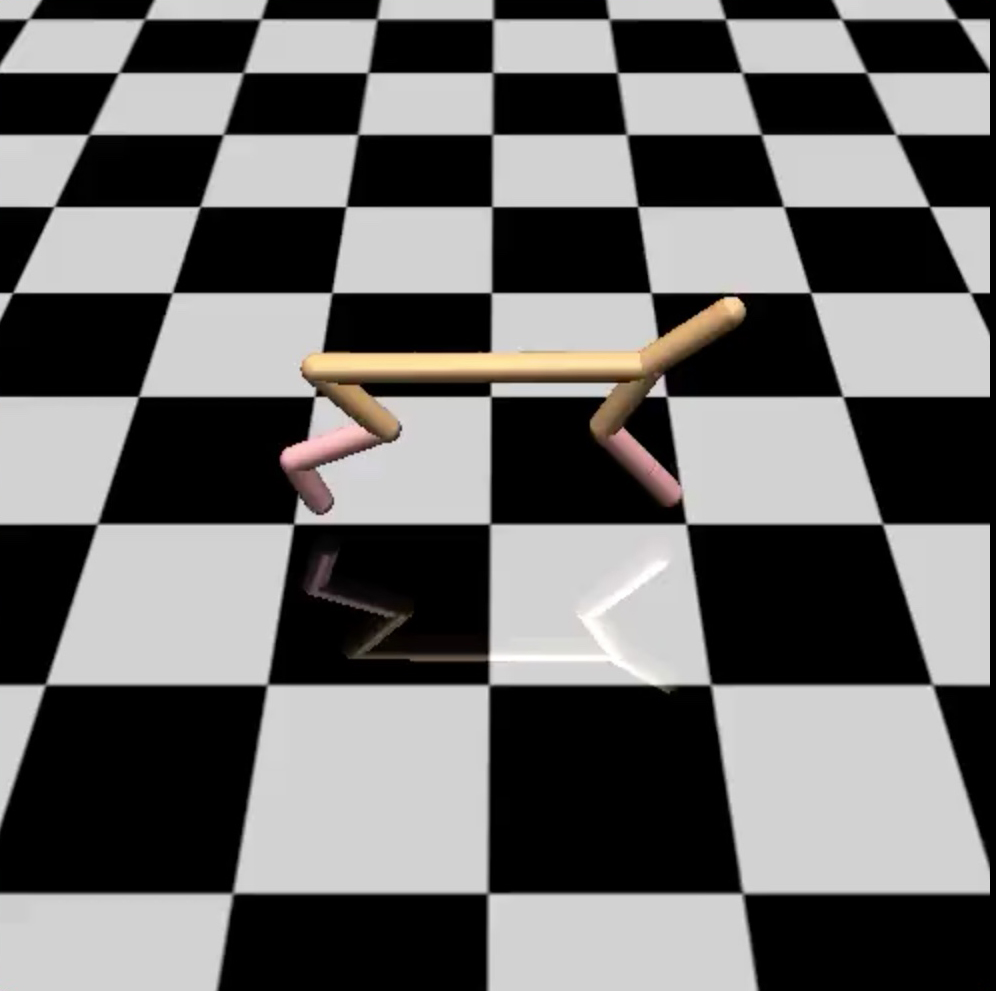}
    \caption{The MuJoCo HalfCheetah domain.}
    \label{fig:halfhcheetahpic}
\end{figure}

As stated earlier, in this experiment, we aim to demonstrate the superiority of the doubt rule over the discrepancy rule in a more complex domain with a large state and action space. Further, we show the effect of combining instances of the doubt and discrepancy rule under the EnsembleDAgger decision rule. We find that the resulting decision rule is strictly more conservative, but in some cases can lead to improvement over the component decision rules. Decision rules are compared across various settings of the hyperparameters $\chi$ and $\tau$, which are selected \textit{a priori} and held constant over a given experiment, unlike in the previous experiment.

The MuJoCo HalfCheetah-v1 domain (shown in \Cref{fig:halfhcheetahpic}) is an OpenAI Gym environment with observations in $\mathbb{R}^{18}$ and actions in $\mathbb{R}^6$~\cite{openaigym}. We train an expert policy on this domain using the TRPO algorithm from the rllab codebase~\cite{rllab}. The goal is to learn a stable gait, with a reward for the distance from the origin reached. The purpose of this experiment is to compare the doubt rule, discrepancy rule, and a combination of the rules, in their ability to safely learn a policy that matches the expert score.

In this experiment, we use an ensemble of five neural networks, each with five hidden layers with 16-neuron widths, as the policy being trained. 
We use a smaller ensemble than in the last experiment to reduce simulation time. In general, the computational complexity of querying and training the novice policy is linear in the ensemble size, though larger ensembles give more accurate estimates of novice doubt.

The system is trained for seven epochs. Each epoch samples one additional trajectory of interaction with the environment, followed by re-training the novice policy on the aggregated dataset. Each trajectory is an episode with a maximum length of 100 time-steps. When training the policy on the aggregated dataset, we use a learning rate of $10^{-3}$, a batch size of 32, and train for 2000 optimization epochs. Since the expert policy is stochastic, it is appropriate for the neural networks to predict the parameters of a Gaussian distribution as opposed to a point estimate of the actions. 
However, the networks used in this experiment simply predict point estimates since they are easier to train, and epistemic uncertainty is still captured. When training, the score over a trajectory of the lone novice and the system combined under the experiment's decision rule are queried at each epoch. Queries average the score of the policy being tested over 20 trajectories.

We define the \textit{performance} of an instance of a given decision rule by the performance of the lone novice and the performance of the combined system, which are defined as follows. The performance of the lone novice is the average score of the novice, trained under a given decision rule instance, summed over the seven epochs of training. A better performing lone novice implies that the decision rule instance is able to quickly bring the novice to expert-level scores. Similarly, the performance of the combined system is the average score of the expert and novice, combined under a given decision rule instance, summed over the seven epochs that train the novice. 

Though we have no strict notion of safety in this domain, a better performing combined system implies that trajectories perturbed under the decision rule instance are still high-scoring, are thus compromising states that one may consider to be failure states are being better avoided. We compare decision rules based on their ability to maximize the novice performance while compromising the performance of the combined system as little as possible. We sample the performance of each instance of a decision rule 100 times, presenting their mean and standard errors.

We test the doubt rule and the discrepancy rule using values of: %
\begin{equation*}
\begin{array}{rl}
    \chi &= [0.02, 0.05,0.1,0.2,0.5] \\
    \tau &= [0.2,0.5,1.0,2.0,5.0], 
\end{array}
\end{equation*}
and the full EnsembleDAgger decision rule at $(\tau, \chi) = [(0.2,0.02), (0.5,0.05), (1.0,0.1), (2.0,0.2), (5.0,0.5)].$ 

Though we sample $(\tau, \chi)$ only on a line in the positive quadrant of $\mathbb{R}^2$, this ratio between $\tau$ and $\chi$ is chosen so the two rules are approximately equally responsible for preventing the novice action in the first training epoch. 

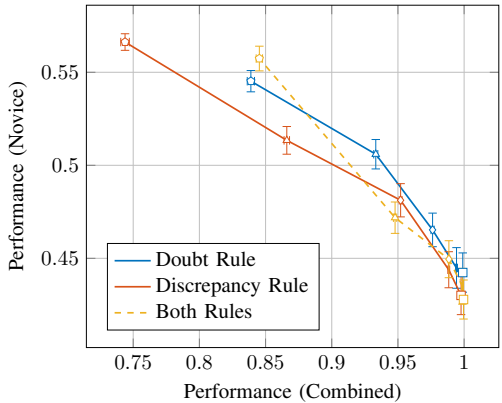
\begin{figure}
		\centering
		\scalebox{0.8}{\begin{tikzpicture}[]
\begin{axis}[legend style = {{at={(0.05,0.05)},anchor=south west}}, ylabel = {Performance (Novice)}, xlabel = {Performance (Combined)}, grid=both]\addplot+ [
C1, solid, thick, no marks, error bars/.cd, 
error bar style = {solid}, x dir=both, x explicit, y dir=both, y explicit]
table [
x error plus=ex+, x error minus=ex-, y error plus=ey+, y error minus=ey-
] {
x y ex+ ex- ey+ ey-
0.9989671 0.44234717 0.00060018344 0.00060018344 0.010556282 0.010556282
0.9942579 0.44483092 0.0007155957 0.0007155957 0.010924337 0.010924337
0.97617733 0.46527392 0.0012243161 0.0012243161 0.009025922 0.009025922
0.9332595 0.50595486 0.001547466 0.001547466 0.0079082325 0.0079082325
0.8388986 0.5452217 0.0027453941 0.0027453941 0.00572594 0.00572594
};
\addlegendentry{Doubt Rule}
\addplot+ [
C2, solid, thick, no marks, error bars/.cd, 
error bar style = {solid}, x dir=both, x explicit, y dir=both, y explicit]
table [
x error plus=ex+, x error minus=ex-, y error plus=ey+, y error minus=ey-
] {
x y ex+ ex- ey+ ey-
0.9976788 0.43007824 0.00062634796 0.00062634796 0.010348183 0.010348183
0.98838145 0.4438655 0.0007087399 0.0007087399 0.009689852 0.009689852
0.9520279 0.4812027 0.0010895671 0.0010895671 0.008932637 0.008932637
0.8660381 0.51340663 0.0020624434 0.0020624434 0.00744242 0.00744242
0.74383587 0.56625676 0.0032423912 0.0032423912 0.004458264 0.004458264
};
\addlegendentry{Discrepancy Rule}
\addplot+ [
C3, dashed, thick, no marks, error bars/.cd, 
error bar style = {solid}, x dir=both, x explicit, y dir=both, y explicit]
table [
x error plus=ex+, x error minus=ex-, y error plus=ey+, y error minus=ey-
] {
x y ex+ ex- ey+ ey-
0.99955124 0.42785567 0.0006837687 0.0006837687 0.0104937935 0.0104937935
0.99872386 0.43496284 0.000635221 0.000635221 0.009621962 0.009621962
0.9884894 0.44951972 0.0006636592 0.0006636592 0.009940047 0.009940047
0.9478382 0.47183985 0.0012502901 0.0012502901 0.00843376 0.00843376
0.8452989 0.557391 0.0024140542 0.0024140542 0.006634201 0.006634201
};
\addlegendentry{Both Rules}
\addplot+ [mark=square*, C1, solid, mark options={fill=white}]coordinates {
(0.9989671, 0.44234717)
};
\addplot+ [mark=square*, C2, solid, mark options={fill=white}]coordinates {
(0.9976788, 0.43007824)
};
\addplot+ [mark=square*, C3, solid, mark options={fill=white}]coordinates {
(0.99955124, 0.42785567)
};
\addplot+ [mark=circle*, C1, solid, mark options={fill=white}]coordinates {
(0.9942579, 0.44483092)
};
\addplot+ [mark=circle*, C2, solid, mark options={fill=white}]coordinates {
(0.98838145, 0.4438655)
};
\addplot+ [mark=circle*, C3, solid, mark options={fill=white}]coordinates {
(0.99872386, 0.43496284)
};
\addplot+ [mark=diamond*, C1, solid, mark options={fill=white}]coordinates {
(0.97617733, 0.46527392)
};
\addplot+ [mark=diamond*, C2, solid, mark options={fill=white}]coordinates {
(0.9520279, 0.4812027)
};
\addplot+ [mark=diamond*, C3, solid, mark options={fill=white}]coordinates {
(0.9884894, 0.44951972)
};
\addplot+ [mark=triangle*, C1, solid, mark options={fill=white}]coordinates {
(0.9332595, 0.50595486)
};
\addplot+ [mark=triangle*, C2, solid, mark options={fill=white}]coordinates {
(0.8660381, 0.51340663)
};
\addplot+ [mark=triangle*, C3, solid, mark options={fill=white}]coordinates {
(0.9478382, 0.47183985)
};
\addplot+ [mark=pentagon*, C1, solid, mark options={fill=white}]coordinates {
(0.8388986, 0.5452217)
};
\addplot+ [mark=pentagon*, C2, solid, mark options={fill=white}]coordinates {
(0.74383587, 0.56625676)
};
\addplot+ [mark=pentagon*, C3, solid, mark options={fill=white}]coordinates {
(0.8452989, 0.557391)
};
\end{axis}

\end{tikzpicture}}
		\caption{\small Performance of various instances of the doubt rule, discrepancy rule, and combined EnsembleDAgger decision rules on the HalfCheetah domain. Consistent markers indicate the instances of the doubt and discrepancy ruled in an instance of the EnsembleDAgger decision rule. \label{fig:halfcheetah}}
	\end{figure}

The performance of the various decision rules for the parameters stated are shown in \Cref{fig:halfcheetah} in the form of Pareto frontiers (since varying the hyperparameter trades-off between performance of the combined system and of the lone novice). We see that the doubt rule Pareto dominates the discrepancy rule, as the rule's frontier achieves better novice performance for the same compromise on the combined system's performance. 

The fact that the doubt rule Pareto dominates the discrepancy rule in terms of performance is consistent with the trends observed in the inverted pendulum experiment--the doubt rule constrains the novice to act only when the state is familiar. %
Consequently, the perturbation from an expert action caused by choosing the novice's action is unlikely to compromise the score of the overall trajectory, though it will likely carry the system into marginally more unfamiliar territory, thereby allowing the novice to learn a more robust policy.
The doubt rule, however, allows an arbitrarily large perturbation in sufficiently familiar states, and thereby can still lead to unsafe states. There exist settings of $\chi$ and $\tau$ that can make the EnsembleDAgger decision rule safe in all states, bounding the maximum perturbation from the expert action even in very familiar states. We only sample values of $\tau$ and $\chi$ along a line in $\mathbb{R}^2$, and hence do not find that points along this line show strict improvement over the independent decision rules in all cases, but see slight improvement in novice performance over the doubt rule for the case of $(\tau, \chi) = (0.5, 0.05)$.  Additionally, we find the EnsembleDAgger decision to be strictly more conservative than either of the component decision rules and thus always improves the combined system performance, as expected.

\section{Conclusion}
\label{sec:conclusion}

In this work, we presented an extension to the DAgger algorithm that considers the safety of the novice-expert system that provides the trajectories from which the novice learns. To avoid requiring precise knowledge of safety, we assume the risk of a state to be inversely related to the size of the perturbation to an expert's action that it can accept without compromising safety. We therefore use a model-free proxy for safety by making a key assumption that the risk of a given state correlates with our familiarity with the state in the dataset $\mathcal{D}$. We expect this assumption to hold in domains where the expert is designed to maintain a margin of safety. 
Our algorithm replaces a weighted coin-flip that decides whether the novice acts in the VanillaDAgger decision rule. To act, the novice proposes an action that is bounded in its deviation from the expert's choice of action, as proposed by SafeDAgger*~\cite{Zhang2016}, but also must exhibit low variance in its choice. 

In our experiments, we compared these two conditions independently, calling the first the discrepancy rule and the second the doubt rule. We found that the doubt rule effectively constrains the novice to act only in states it is familiar with, i.e. states that are within some neighborhood of states labeled in $\mathcal{D}$, while the discrepancy rule haphazardly allows the novice to act in states where there is chance agreement between their actions. Since the domain satisfies the assumptions regarding risk that are made, we find that the doubt rule is superior to the discrepancy rule in both its ability to have the novice rapidly attain expert-level control, as well as preventing the novice from carrying the combined expert-novice system into severly compromising states. Though the doubt rule alone is shown to be superior to the discrepancy rule alone, there exist hyperaparameter settings in which the conjunction of the rules is better than either individually.

Future work includes investigating methods for relaxing our risk assumptions, in particular the conflation of safety and familiarity. %
There exist environments with `bottleneck' states in which the expert must frequently travel close to unsafe states to achieve its goal. Additionally, we have not provided a method for choosing the hyperparameters $\chi$ and $\tau$, and thus intend to develop heuristic strategies that can safely discover the most suitable setting of these parameters. A recent work has already demonstrated a method for doing so on real vehicles~\cite{kelly2018hg}.

\section*{Acknowledgments}
This material is based upon work supported by SAIC Innovation Center, a subsidiary of SAIC Motors. The authors would like to acknowledge the useful feedback of Apoorva Sharma and Michael Kelly.

\newpage
\bibliographystyle{IEEEtran}
\bibliography{references}

\newpage
\appendix
\section{Appendix}

\subsection{Bayesian Approximation Techniques}
\label{app:GP_approx}

We examine two approximations of GPs: ensemble methods and Monte-Carlo dropout.

Gal et al. propose approximating Bayesian models with neural networks trained with dropout~\cite{Gal2015DropoutB}. %
By applying dropout at every weight layer in a network, an approximation of a Gaussian process is obtained. 
Given a policy trained with dropout, the network can be queried $N$ times per input observation to obtain a distribution over actions, using randomly sampled dropout masks~\cite{Gal2015DropoutB,srivastava2014dropout}.

The ensemble method is a technique for training a collection of neural networks to execute the same task and then combining the output into a single prediction.
This approach has shown to significantly improve performance in practice~\cite{zhou2002ensembling}.
\cite{lakshminarayanan2017simple} employed an ensemble of neural networks to approximate GPs and demonstrated that this is a more straightforward approach to estimate predictive uncertainty (PU). 
Typically, neural networks predict point estimates of the output are optimized to minimize the mean squared error on the training set. 
The authors claim that this does not capture \textit{aleatoric} uncertainty, but only \textit{epistemic} uncertainty. They propose using a \emph{proper scoring rule}, like negative log-likelihood, as a loss function to train an ensemble in which each network predicts a mean and a variance of a Gaussian distribution over the output.
They postulate that such loss functions provide a better measure of the quality of predictive uncertainty and thus reward better calibrated predictions.

\subsubsection{Empirical Evaluation}
To determine which approximation approach is most effective in practice, we evaluate the ability of four different methods to learn the function $f(x) = \sin(\pi x) + 0.2\sin(4\pi x)$ with only eight samples. 
The codebase used for the evaluation (as well as the proposed algorithms) is provided in the supplementary material.

First, we fit the function with a traditional Gaussian process to act as a baseline.
We use a squared-exponential kernel with a length-scale of 10. The kernel parameters chosen are the best of nine optimizer restarts~\cite{scikit-learn}.
Then, we fit a ten network ensemble trained with Mean-Square-Error (MSE) loss for 300 epochs, referred to as the Vanilla Ensemble. 
The same ensemble is trained with Negative-Log-Likelihood (NLL) loss for 2400 epochs, so each network directly predicts uncertainty.
Finally, a single network is trained using Monte-Carlo Dropout with MSE loss and a keep-probability of 75\%, also trained for 2400 epochs.

All neural network models have hidden layers of size $[128, 64, 64, 64]$, respectively. ADAM with a learning rate of $10^{-3}$ is used to optimize the vanilla ensemble and MC-dropout, while a learning rate of $10^{-4}$ is used for the ensemble with predictive uncertainty. No weight regularization or batch normalization are used. We use a batch size of 4.

\begin{figure}[h!]
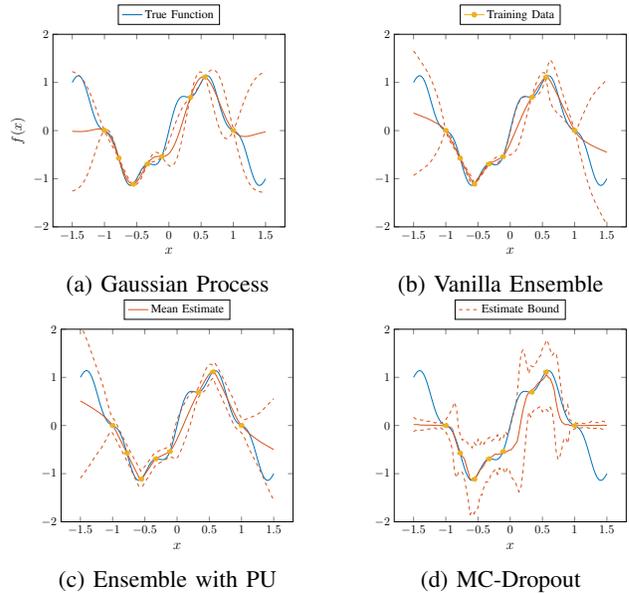

	\begin{subfigure}{0.24\textwidth}
		\scalebox{0.45}{\input{Figures/gp_pgf}}
		\caption{Gaussian Process}
	\end{subfigure}~
	\begin{subfigure}{0.24\textwidth}
		\centering
		\scalebox{0.45}{\input{Figures/mse_pgf}}
		\caption{Vanilla Ensemble}
	\end{subfigure} \\
	\begin{subfigure}{0.24\textwidth}
		\centering
		\scalebox{0.45}{\input{Figures/nll_pgf}}
		\caption{Ensemble with PU}
	\end{subfigure}~
	\begin{subfigure}{0.24\textwidth}
		\centering
		\scalebox{0.45}{\input{Figures/drop_pgf}}
		\caption{MC-Dropout}
	\end{subfigure}
	\caption{\small Comparing Gaussian process approximation methods.}
	\label{fig:gp_comparison}
\end{figure}

Each model is queried for a mean and standard deviation for its estimate of $f(x)$ for $x \in [-1.5,1.5]$. 
The standard deviations of each model are scaled such that their sum matches that of the GP, for ease of visual comparison. 
The results of each method are shown in \Cref{fig:gp_comparison}.

We see that the vanilla ensemble of models and ensemble with predictive uncertainty have the most visual similarity to a GP.
We also note that the vanilla ensemble achieves this performance in a small fraction of the number of epochs with which the latter two models are trained.
Therefore, in our experiments, the novice neural network architecture takes the form of an ensemble of neural networks, and we train the neural network with MSE loss. %

It should be noted that the implementations did not utilize adversarial training, as recommended by Lakshminarayanan et al.~\cite{lakshminarayanan2017simple}. Additionally, the training data is noiseless, which does not highlight the potential benefit of using an ensemble with predictive uncertainty or MC-Dropout.

\begin{figure*}[t]
	\centering
	\begin{subfigure}{0.32\textwidth}
		\centering
		\scalebox{0.55}{\begin{tikzpicture}[]
\begin{axis}[legend style = {{at={(0.25,0.3)}, anchor=north west}}, ylabel = {Learning Performance}, xlabel = {Epoch}, grid=both]\addplot+ [
C1, mark options={fill=C1}, error bars/.cd, 
x dir=both, x explicit, y dir=both, y explicit]
table [
x error plus=ex+, x error minus=ex-, y error plus=ey+, y error minus=ey-
] {
x y ex+ ex- ey+ ey-
1.0 0.022777779 0.0 0.0 0.007412274833768606 0.007412274833768606
2.0 0.41685185 0.0 0.0 0.046555452048778534 0.046555452048778534
3.0 0.65277773 0.0 0.0 0.05298641324043274 0.05298641324043274
4.0 0.777037 0.0 0.0 0.0469721294939518 0.0469721294939518
5.0 0.84722227 0.0 0.0 0.04301150143146515 0.04301150143146515
6.0 0.87203705 0.0 0.0 0.03899318352341652 0.03899318352341652
};
\addlegendentry{Doubt Rule $\chi =10^{-3}$}
\addplot+ [
C2, mark options={fill=C2}, error bars/.cd, 
x dir=both, x explicit, y dir=both, y explicit]
table [
x error plus=ex+, x error minus=ex-, y error plus=ey+, y error minus=ey-
] {
x y ex+ ex- ey+ ey-
1.0 0.01740741 0.0 0.0 0.0058402977883815765 0.0058402977883815765
2.0 0.39074075 0.0 0.0 0.04832833260297775 0.04832833260297775
3.0 0.53166664 0.0 0.0 0.042548272758722305 0.042548272758722305
4.0 0.57611114 0.0 0.0 0.043337222188711166 0.043337222188711166
5.0 0.5914815 0.0 0.0 0.04312022402882576 0.04312022402882576
6.0 0.64277774 0.0 0.0 0.04574808105826378 0.04574808105826378
};
\addlegendentry{Discrep. Rule $\tau =10^{-1}$}
\addplot+ [
C3, mark options={fill=C3}, error bars/.cd, 
x dir=both, x explicit, y dir=both, y explicit]
table [
x error plus=ex+, x error minus=ex-, y error plus=ey+, y error minus=ey-
] {
x y ex+ ex- ey+ ey-
1.0 0.04777778 0.0 0.0 0.014395239762961864 0.014395239762961864
2.0 0.32166666 0.0 0.0 0.047465451061725616 0.047465451061725616
3.0 0.41500002 0.0 0.0 0.054375823587179184 0.054375823587179184
4.0 0.46759257 0.0 0.0 0.056093234568834305 0.056093234568834305
5.0 0.6005556 0.0 0.0 0.06231164187192917 0.06231164187192917
6.0 0.6394445 0.0 0.0 0.059566184878349304 0.059566184878349304
};
\addlegendentry{Discrep. Rule $\tau =5\cdot 10^{-2}$}
\end{axis}

\end{tikzpicture}}
		\caption{Learning Performance \label{fig:bothfixedlearning}}
	\end{subfigure}~
	\begin{subfigure}{0.32\textwidth}
		\centering
		\scalebox{0.55}{\begin{tikzpicture}[]
\begin{axis}[ylabel = {Failure Rate [$\%$]}, xlabel = {Epoch}, grid=both]\addplot+ [C1, mark options={fill=C1}]coordinates {
(1.0, 0.0)
(2.0, 0.0)
(3.0, 0.0)
(4.0, 0.0)
(5.0, 0.0)
(6.0, 0.0)
};
\addplot+ [C2, mark options={fill=C2}]coordinates {
(1.0, 3.3333335)
(2.0, 10.0)
(3.0, 10.0)
(4.0, 6.666667)
(5.0, 6.666667)
(6.0, 3.3333335)
};
\addplot+ [C3, mark options={fill=C3}]coordinates {
(1.0, 0.0)
(2.0, 0.0)
(3.0, 0.0)
(4.0, 3.3333335)
(5.0, 0.0)
(6.0, 0.0)
};
\end{axis}

\end{tikzpicture}}
		\caption{Failure Rate \label{fig:bothfixedfailures}}
	\end{subfigure}~
	\begin{subfigure}{0.32\textwidth}
		\centering
		\scalebox{0.55}{\begin{tikzpicture}[]
\begin{axis}[ylabel = {Permitted Set Volume}, xlabel = {Epoch}, grid=both]\addplot+ [
C1, mark options={fill=C1}, error bars/.cd, 
x dir=both, x explicit, y dir=both, y explicit]
table [
x error plus=ex+, x error minus=ex-, y error plus=ey+, y error minus=ey-
] {
x y ex+ ex- ey+ ey-
1.0 0.028666664 0.0 0.0 0.00211973930709064 0.00211973930709064
2.0 0.077999994 0.0 0.0 0.00731024332344532 0.00731024332344532
3.0 0.097666666 0.0 0.0 0.006929378490895033 0.006929378490895033
4.0 0.10966667 0.0 0.0 0.006547498516738415 0.006547498516738415
5.0 0.12366667 0.0 0.0 0.005730706267058849 0.005730706267058849
6.0 0.13833334 0.0 0.0 0.004728754051029682 0.004728754051029682
};
\addplot+ [
C2, mark options={fill=C2}, error bars/.cd, 
x dir=both, x explicit, y dir=both, y explicit]
table [
x error plus=ex+, x error minus=ex-, y error plus=ey+, y error minus=ey-
] {
x y ex+ ex- ey+ ey-
1.0 0.27341664 0.0 0.0 0.013849292881786823 0.013849292881786823
2.0 0.39150003 0.0 0.0 0.016756195574998856 0.016756195574998856
3.0 0.42241663 0.0 0.0 0.010842050425708294 0.010842050425708294
4.0 0.43725002 0.0 0.0 0.008907531388103962 0.008907531388103962
5.0 0.44149998 0.0 0.0 0.007671974133700132 0.007671974133700132
6.0 0.4455 0.0 0.0 0.0073072942905128 0.0073072942905128
};
\addplot+ [
C3, mark options={fill=C3}, error bars/.cd, 
x dir=both, x explicit, y dir=both, y explicit]
table [
x error plus=ex+, x error minus=ex-, y error plus=ey+, y error minus=ey-
] {
x y ex+ ex- ey+ ey-
1.0 0.18616667 0.0 0.0 0.009720765985548496 0.009720765985548496
2.0 0.34 0.0 0.0 0.021376147866249084 0.021376147866249084
3.0 0.36991665 0.0 0.0 0.015576324425637722 0.015576324425637722
4.0 0.38499996 0.0 0.0 0.012683708220720291 0.012683708220720291
5.0 0.38175 0.0 0.0 0.011222763918340206 0.011222763918340206
6.0 0.38308334 0.0 0.0 0.0074023171328008175 0.0074023171328008175
};
\end{axis}

\end{tikzpicture}}
		\caption{Permitted Set Volume \label{fig:bothfixedareas}}
	\end{subfigure}
	
	\caption{\small Comparison of the learning performance, failure rate, and permitted set volume for the doubt and discrepancy rule, where $\chi$ and $\tau$ are chosen \emph{a priori}.%
	}
	\label{fig:bothfixed}
	\vspace{-10pt}
\end{figure*}
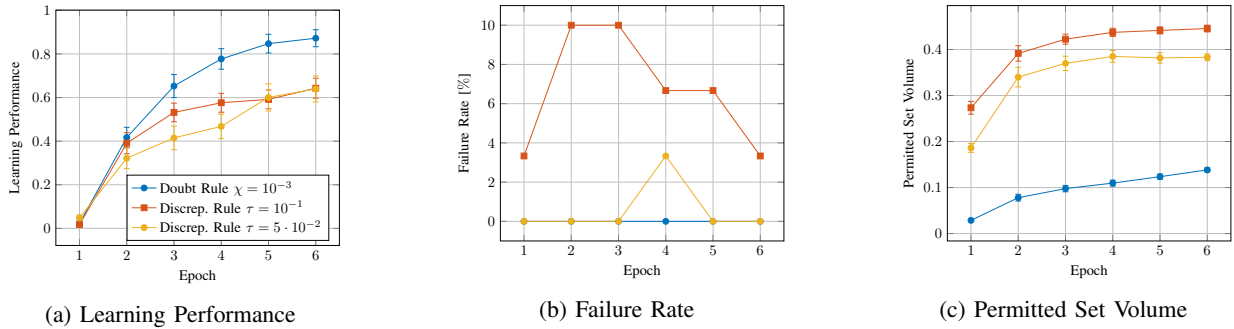

\subsection{The Inverted Pendulum domain}
\label{app:invpen}
The inverted pendulum has a two-dimensional state space of $[\theta,\dot{\theta}]$ and a one-dimensional action space of $u$, as shown in~\Cref{fig:invpen_diagram}.

The inverted pendulum is guided by the following dynamical equation: %
\begin{align}
	\ddot{\theta} = \frac1a\sin\theta - b\dot{\theta} + cu
\end{align}
The system can be driven to the equilibrium at $[0,0]$ by feedback linearization, using the following control law:
\begin{align}
	u = -\frac{a}{c}\sin\theta - 
	\frac1c \begin{bmatrix}K_1 & K_2\end{bmatrix}\begin{bmatrix}\theta \\ \dot{\theta}\end{bmatrix}
\end{align}
With feedback linearization, the gain vector $K$ is computed to stabilize the linear system specified by the new dynamics:
\begin{align}
	\begin{bmatrix} \dot{v_1} \\ \dot{v_2} \end{bmatrix} = 
	\begin{bmatrix} 0 & 1 \\ 0 & -b \end{bmatrix} \begin{bmatrix}v_1 \\ v_2 \end{bmatrix} + 
	\begin{bmatrix}0\\1 \end{bmatrix} u
\end{align}
where $v_1$ and $v_2$ are the residual linear terms. The resulting controller is deterministic but sufficiently non-linear to pose an interesting learning problem. 
The dynamics of this environment and control law found by feedback linearization follow the example presented by Khalil et al.~\cite{khalilnonlinear}.

We consider the problem instance in which $a = 10$, $b = 2$, $c = 10$. Additionally, the control $u$ is saturated to lie within $[-1,1]$. The gains $K$ are found to be $[0.316, 0.175]$ using a linear quadratic regulator with cost function $J = \int_0^\infty v_1^2 + v_2^2 + 10u^2$. Due to the control saturation, the controller does not converge to the desired fixed point from an arbitrary initial condition, but has the basin of attraction shown in \Cref{fig:dom_attraction}. \Cref{fig:dom_attraction} also shows the region of the state space from which initial conditions are sampled uniformly during the successive epochs of DAgger.

\subsection{Inverted Pendulum: Selecting Hyperparameters} %
\label{app:invpenhyp}

In the inverted pendulum experiment discussed in the body of this paper, we solve for the hyperparameters $\chi$ and $\tau$ at every DAgger epoch such that the corresponding decision rules create a permitted set of some desired volume. Though useful for visually comparing the two decision rules, solving for the hyperparameters in this manner is not tractable in more complex problems with higher dimensional state-action spaces or with a non-deterministic expert. Hence, in our second experiment, we compare the behavior of the two decision rules in a manner more useful to a practitioner---in which the hyperparameters $\chi$ and $\tau$ are chosen \textit{a priori}.

We introduce additional metrics of performance: 
\begin{itemize}
	\item \textit{Learning Performance}: The fraction of states grid-sampled from $\theta \in [-\pi,\pi], \dot{\theta} \in [-5,5]$,  in both the expert and  novice basin of attractions. 
	\item \textit{Failure Rate}: The fraction of repetitions of a given experiment in which the trajectory acquired at a given epoch, for a given decision rule and expert policy, leaves the expert's basin of attraction.
\end{itemize}
For a given choice of hyperparameters, the instances of the decision rules are compared over six epochs, and results are averaged over 30 repetitions of the experiment. During each epoch, we track the learning performance, failure rate, and permitted set volume.

\Cref{fig:bothfixed} shows the results for an instance of the doubt rule with $\chi=10^{-3}$, the discrepancy rule with $\tau=10^{-1}$ and the discrepancy rule with $\tau=5\cdot10^{-2}$. As shown in \Cref{fig:bothfixedlearning}, this instance of the doubt rule demonstrates superior learning performance to either instance of the discrepancy rule. In addition to demonstrating more rapid learning, this instance exhibits no failures in any of the six epochs in any of the repeated experiments, as shown in \Cref{fig:bothfixedfailures}. Neither instance of the discrepancy rule is failure-free, and choosing the more conservative $\tau$ reduces the failure rate at the expense of learning performance.

It is also interesting to note the evolution of the permitted set volume over the six epochs for fixed hyperparameter choices. It appears that in this domain, the permitted set volumes grow monotonically, which matches expectations. However, we can see that the permitted set volumes for the discrepancy rule are many times larger than that of the doubt rule. This confirms the observations made in the previous experiment: the doubt rule is less permissive in allowing the novice to act, but is nonetheless able to generative more informative trajectories.

The results of this experiment confirm the observations made in the first inverted pendulum quantitatively. As a consequence of the discrepancy rule being too conservative in states familiar to the novice, the system is prevented from entering states that are informative to the novice's learning, and hence we see poor learning performance. Furthermore, since the discrepancy rule is not conservative enough in risky states, the discrepancy rule encounters failures more frequently than the doubt rule.

\end{document}